\documentclass[11pt,a4paper]{article}
\usepackage[preprint]{acl}
\usepackage{times}
\usepackage{latexsym}
\usepackage{url}
\usepackage{graphicx}
\usepackage{booktabs}
\usepackage{amsmath}
\usepackage{amssymb}
\usepackage{multirow}
\usepackage{tikz}
\usetikzlibrary{arrows.meta, positioning,calc}
\usetikzlibrary{shapes}
\usepackage{ifthen}
\usepackage{subcaption}
\usepackage{CJKutf8}
\usepackage[most]{tcolorbox}

\newtcolorbox{promptbox}[1]{
  breakable,
  colback=orange!15!white,
  colframe=orange!70!black,
  coltext=black,
  boxrule=0.5mm,
  title={\textbf{#1}},
  fonttitle=\bfseries,
}

\title{Semantic Outlier Removal with Embedding Models and LLMs}

\author{
  \textbf{Eren Akbiyik}, 
  \textbf{Jo{\~a}o Almeida}, 
  \textbf{Rik Melis},\\
  \textbf{Ritu Sriram}, 
  \textbf{Viviana Petrescu}, 
  \textbf{Vilhj{\'a}lmur Vilhj{\'a}lmsson}
\\[1ex]
TripleLift\\
Zürich, Switzerland
\\[1ex]
  \small{
    \textbf{Correspondence:} \href{mailto:eakbiyik@triplelift.com}{eakbiyik@triplelift.com}
  }
}

\begin{document}
\maketitle
\begin{abstract}
Modern text processing pipelines demand robust methods to remove extraneous content while preserving a document's core message. Traditional approaches—such as HTML boilerplate extraction or keyword filters—often fail in multilingual settings and struggle with context-sensitive nuances, whereas Large Language Models (LLMs) offer improved quality at high computational cost. We introduce SORE (Semantic Outlier Removal), a cost-effective, transparent method that leverages multilingual sentence embeddings and approximate nearest-neighbor search to identify and excise unwanted text segments. By first identifying core content via metadata embedding and then flagging segments that either closely match predefined outlier groups or deviate significantly from the core, SORE achieves near-LLM extraction precision at a fraction of the cost. Experiments on HTML datasets demonstrate that SORE outperforms structural methods and yield high precision in diverse scenarios. Our system is currently deployed in production, processing millions of documents daily across multiple languages while maintaining both efficiency and accuracy. To facilitate further research, we will publicly release our implementation and evaluation datasets.
\end{abstract}

\section{Introduction}
Effective content extraction from web pages is a critical component in many modern NLP pipelines, enabling cleaner inputs for downstream tasks such as summarization, classification, and information retrieval. However, web documents typically contain significant amounts of extraneous content—navigation elements, advertisements, legal disclaimers, related article recommendations, and other non-essential text—that can degrade the performance of these tasks.

Traditional approaches to this problem include HTML-structure-based methods like Readability.js \cite{readabilityjs} and Boilerpipe \cite{kohlschutter2010boilerpipe}, which leverage DOM and formatting patterns to identify main content. While efficient, these methods often fail when faced with diverse HTML structures, especially across multiple languages and website designs. They also struggle to distinguish semantically irrelevant text that shares structural characteristics with the main content.

More recently, Large Language Models (LLMs) have demonstrated impressive capabilities in content extraction \cite{brown2020language}, as they can understand the semantic meaning and context of text. However, deploying LLMs at scale incurs substantial computational costs, introducing latency and budget concerns for production systems processing millions of documents. Additionally, LLMs may introduce hallucinations or unpredictable behaviors that compromise reliability.

To address these limitations, we introduce SORE (Semantic Outlier Removal), a system that bridges the gap between traditional structure-based methods and LLMs by utilizing multilingual embedding models. SORE leverages semantic similarity to identify core content by measuring similarity to document metadata, detect outliers by measuring distance to predefined outlier categories, and remove unwanted content while providing transparent justification.

Our approach offers several key advantages for industrial applications. First, SORE operates in a language-agnostic manner, enabling effective content extraction across diverse languages without requiring language-specific rules. Second, it provides transparency with clear explanations for why specific text segments are removed, facilitating debugging and continuous improvement. Third, SORE achieves near-LLM quality extraction at a fraction of the computational cost—a critical factor for production systems processing millions of documents. Finally, its implementation using approximate nearest neighbor search ensures scalability even with large document volumes.

This paper describes the SORE algorithm, its implementation details optimized for production deployment, and comprehensive experiments demonstrating its effectiveness compared to both traditional methods and LLM-based approaches. We also provide a detailed cost analysis, highlighting the significant efficiency gains achieved by our approach. To promote reproducibility and facilitate further research, we will make our implementation and evaluation datasets publicly available.

\section{Related Work}
\subsection{HTML Boilerplate Removal}
Extracting main content from HTML documents remains challenging in web information retrieval. \citet{kohlschutter2010boilerpipe} introduced text density features to identify boilerplate content, while Readability.js \cite{readabilityjs} employs heuristic rules based on HTML structure. Despite their efficiency, these approaches struggle with complex layouts and multilingual content.

\subsection{Embedding Models for Text Similarity}
Dense vector representations have transformed NLP by capturing semantic relationships between texts. Evolving from word embeddings \cite{mikolov2013distributed,pennington2014glove} to sentence representations, models like Sentence-BERT \cite{reimers2019sentence} adapted transformer architectures for similarity tasks. Multilingual embedding models \cite{artetxe2019massively, wang2024multilingual} now enable cross-lingual applications, with commercial services like Cohere \cite{cohere} and AWS Titan offering production-ready solutions.

\subsection{LLMs for Content Extraction}
LLMs demonstrate strong capabilities in understanding contextual meaning \cite{brown2020language, scao2023bloom}, making them promising for content extraction. However, they require significant computational resources and may produce inconsistent outputs \cite{bender2021dangers}. Their effectiveness varies across languages, particularly for lower-resource ones \cite{nguyen2023culturax}.

\subsection{Outlier Detection in Text}
Text outlier detection approaches include density-based methods \cite{sereshki2023textual} and embedding space analysis \cite{haemmerl2023exploring}. Most work focuses on document-level detection rather than identifying outlier segments within documents.

Our work bridges these areas by leveraging embedding-based similarity with efficient nearest-neighbor search for multilingual outlier content identification, balancing traditional methods' efficiency with LLMs' semantic understanding.

\section{SORE: System Design and Implementation}
We introduce SORE (Semantic Outlier Removal), a method for removing unwanted text segments from documents based on semantic similarity. SORE identifies and removes text segments that match known patterns of boilerplate content or semantically diverge from the document's theme.

\newcommand{\PlaceSentenceNodes}[1]{%
  \ifthenelse{\equal{#1}{true}}%
    {\def\mylabel##1{\tiny ##1}}
    {\def\mylabel##1{}}

  \node[circle, fill=black!15, draw, minimum size=0.5cm, label=\mylabel{S1}] (s1) at (0.3,1.3) {};
  \node[circle, fill=black!15, draw, minimum size=0.5cm, label=\mylabel{S2}] (s2) at (1.2,0.4) {};
  \node[circle, fill=black!15, draw, minimum size=0.5cm, label=\mylabel{S3}] (s3) at (1.5,1.6) {};
  \node[circle, fill=black!15, draw, minimum size=0.5cm, label=\mylabel{S4}] (s4) at (1.2,-0.9) {};
  \node[circle, fill=black!15, draw, minimum size=0.5cm, label=\mylabel{S5}] (s5) at (-1.1,-1.0) {};
  \node[circle, fill=black!15, draw, minimum size=0.5cm, label=\mylabel{S6}] (s6) at (-1.2,1.6) {};
}

\begin{figure*}[t]
\centering

\begin{subfigure}[t]{0.24\textwidth}
\centering
\begin{tikzpicture}[font=\sffamily, baseline=(current bounding box.north), scale=0.7]

\node[anchor=north, align=center, yshift=0.3cm, font=\small]
  at (-1.5,2.6) {(a) Segmentation \& Embedding};

\node[draw, fill=gray!10, text width=1.2cm, minimum height=1.2cm, align=left] (html) 
  at (-3,0)
  {\scriptsize
   \texttt{<html>}\\
   Title\\
   Body...\\
   Cookie...\\
   Footer...\\
   \texttt{</html>}
  };

\draw[->, thick] (html.east) -- (-1.0,0);

\node[circle, fill=orange!20, draw, minimum size=0.7cm] (meta) at (0,0) {\tiny $w_m$};

\PlaceSentenceNodes{true}

\end{tikzpicture}
\end{subfigure}
\hspace{0.5cm}
\hfill
\begin{subfigure}[t]{0.22\textwidth}
\centering
\begin{tikzpicture}[font=\sffamily, baseline=(current bounding box.north), scale=0.7]

\node[anchor=north, align=center, yshift=0.3cm, font=\small]
  at (0,2.6) {(b) Core Detection};

\node[circle, fill=orange!40, draw, thick, minimum size=0.7cm] (meta) at (0,0) {\tiny $w_m$};

\PlaceSentenceNodes{false}

\foreach \n in {s1,s2,s3,s4,s5,s6} {
  \draw[->, thick, orange!70] (meta) -- (\n);
}

\draw[orange!80, very thick] (s1) circle [radius=0.35];
\draw[orange!80, very thick] (s2) circle [radius=0.35];

\end{tikzpicture}
\end{subfigure}
\hfill
\begin{subfigure}[t]{0.24\textwidth}
\centering
\begin{tikzpicture}[font=\sffamily, baseline=(current bounding box.north), scale=0.7]

\node[anchor=north, align=center, yshift=0.3cm, font=\small]
  at (0,2.6) {(c) Outlier Detection};

\node[circle, fill=orange!40, draw, thick, minimum size=0.7cm] (meta) at (0,0) {\tiny $w_m$};

\PlaceSentenceNodes{false}

\draw[orange!80, very thick] (s1) circle [radius=0.35];
\draw[orange!80, very thick] (s2) circle [radius=0.35];

\draw[dashed, thick, orange!70, rotate around={45:(1.5,0.1)}](0.9,0.9) ellipse (2.5cm and 1.8cm);

\node[circle, fill=purple!30, draw, minimum size=0.4cm, inner sep=1] (o1) at (3.3,1.45) {\tiny Cookie};
\node[circle, fill=purple!30, draw, minimum size=0.4cm, inner sep=1] (o2) at (3.2,0) {\tiny Footer};
\node[circle, fill=purple!30, draw, minimum size=0.4cm, inner sep=1] (o3) at (2.4,-1.2) {\tiny Ads};

\draw[->, thick, orange!80] (s3) -- (s2);
\draw[->, thick, orange!80] (s5) -- (meta);
\draw[->, thick, purple!70] (s4) -- (o3);

\draw[red, very thick] (s6) circle [radius=0.35];
\node[font=\tiny, red, above=0.1 of s6] {};

\draw[red, very thick] (s4) circle [radius=0.35];
\node[font=\tiny, red, above=0.1 of s4] {};

\end{tikzpicture}
\end{subfigure}
\hfill
\begin{subfigure}[t]{0.24\textwidth}
\centering
\begin{tikzpicture}[font=\sffamily, baseline=(current bounding box.north), scale=0.7]

\node[anchor=north, align=center, yshift=0.3cm, font=\small]
  at (0,2.6) {(d) Segment Removal};

\node[circle, fill=orange!40, draw, thick, minimum size=0.7cm] (meta) at (0,0) {\tiny $w_m$};

\PlaceSentenceNodes{false}

\draw[orange!80, very thick] (s1) circle [radius=0.35];
\draw[orange!80, very thick] (s2) circle [radius=0.35];
\draw[orange!80, very thick] (s3) circle [radius=0.35];
\draw[orange!80, very thick] (s5) circle [radius=0.35];

\draw[gray, dashed, thick] (s4) circle [radius=0.45];
\node[font=\tiny, red, below=0.05 of s4] {Removed};

\draw[gray, dashed, thick] (s6) circle [radius=0.45];
\node[font=\tiny, red, below=0.05 of s6] {Removed};

\end{tikzpicture}
\end{subfigure}

\caption{\textbf{Overall pipeline of SORE.}
\textbf{(a) Segmentation \& Embedding:} We split the HTML into segments (S1--S6) and embed them along with a metadata vector $\mathbf{w}_m$. 
\textbf{(b) Core Identification:} Compute similarity of each segment to $\mathbf{w}_m$ and select the top $k\%$ (orange outlines). 
\textbf{(c) Outlier Detection:} Embed predefined outlier groups (purple). For each non-core segment, check distance to the core region (dashed circle) and outlier groups. Flag segments that are too distant from the core or too close to outliers. 
\textbf{(d) Segment Removal:} Remove flagged segments (dashed/gray), keeping the remaining set as the cleaned content.}
\label{fig:teaser}
\vspace{-0.4cm}
\end{figure*}
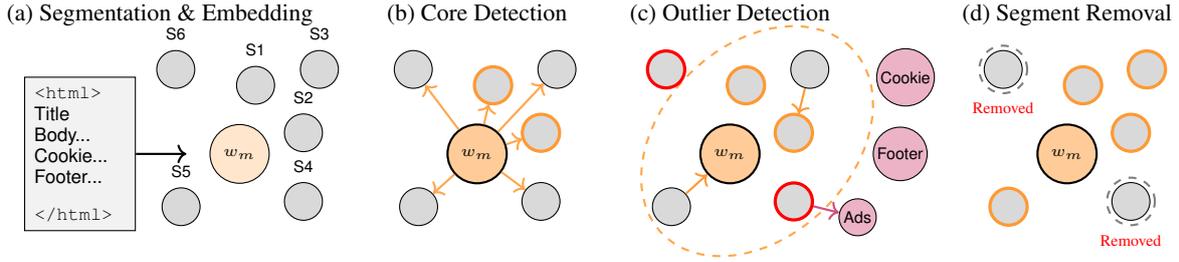

\subsection{Algorithm Overview}
SORE operates through four sequential steps that transform raw HTML content into clean content:

\paragraph{Step 1: Segmentation and Embedding.}
The document is first split into segments (sentences or paragraphs) using an HTML parser that preserves the document structure. Each segment is then converted into a fixed-length dense vector representation using a multilingual embedding model. The document's metadata (e.g., title and description) is also embedded into a vector $\mathbf{w}_m$, which serves as a representation of the document's core theme.

\paragraph{Step 2: Core Identification.}
We compute the cosine distance between each segment's embedding and the metadata embedding $\mathbf{w}_m$. The segments with the smallest distances (highest similarities) to $\mathbf{w}_m$ are selected as the document's "core content". Specifically, we select the top $k\%$ of segments, where $k$ is a configurable parameter that controls the strictness of core content selection.

\paragraph{Step 3: Outlier Detection.}
We define "outlier groups" by embedding phrases indicative of unwanted content types (e.g., advertisements, legal disclaimers, navigation). For each non-core segment, we compute its distance to the closest outlier group and its distance to the core content set. A segment is flagged for removal if either it is too close to an outlier group or it is too distant from the core content (distance above threshold $d$), where $d$ is a configurable distance cutoff parameter.

\paragraph{Step 4: Segment Removal.}
Flagged segments are removed from the document, and the removal reason is recorded (e.g., "matched \textit{disclaimers}" or "too irrelevant"). This explanation provides transparency and aids in system refinement.

\begin{figure}[!tb]
\centering
\begin{tikzpicture}[
    scale=0.8, transform shape,
    node distance=0.8cm,
    box/.style={rectangle, draw, fill=#1, text width=6em, align=center, rounded corners, minimum height=2em},
    arrow/.style={->, thick},
    dasharrow/.style={->, dashed, thick},
    note/.style={font=\scriptsize, align=center, text width=5em}
]

\node[box=blue!10] (input) {HTML Document};
\node[box=blue!10, below=of input] (segment) {Segmented Texts};
\node[box=orange!10, below=of segment] (embed) {Embedding Model};
\node[box=orange!10, below=of embed] (core) {Core Content Identification};
\node[box=orange!10, below=of core] (outlier) {Outlier Detection};
\node[box=blue!10, below=of outlier] (output) {Clean Content};

\node[box=green!10, right=1.6cm of embed] (metadata) {Document Metadata};
\node[box=yellow!10, right=1.6cm of core] (annindex) {ANN Index};
\node[box=green!10, right=1.6cm of outlier] (outgroups) {Outlier Groups};

\draw[arrow] (input) -- node[note, right] {HTML parsing} (segment);
\draw[arrow] (segment) -- (embed);
\draw[arrow] (embed) -- (core);
\draw[arrow] (core) -- (outlier);
\draw[arrow] (outlier) -- node[note, right] {Outlier removal} (output);

\draw[arrow] (metadata) -- (embed);
\draw[arrow] (metadata) -- (core);

\draw[dasharrow] (outgroups) -- (annindex);
\draw[arrow] (annindex) -- (outlier);

\draw[dasharrow] (core) to[bend left=20] (annindex);

\node[note, anchor=west] at ($(outgroups.east) + (-0.25,0)$) {Preloaded at startup};
\node[note, anchor=west, text width=10em] at ($(core)!0.5!(annindex) + (-0.2,0.8)$) {Add core embeddings};
\node[note, anchor=west] at ($(annindex.east) + (0.1,0)$) {HNSW algorithm};

\end{tikzpicture}
\caption{SORE architecture showing the optimized processing pipeline. The system parses HTML documents, segments text, and processes content through an embedding model. Core content is identified using metadata similarity, then an ANN index enables efficient outlier detection by comparing with preloaded outlier groups. This efficient architecture processes millions of documents daily with minimal latency.}
\label{fig:architecture}
\vspace{-0.4cm}
\end{figure}
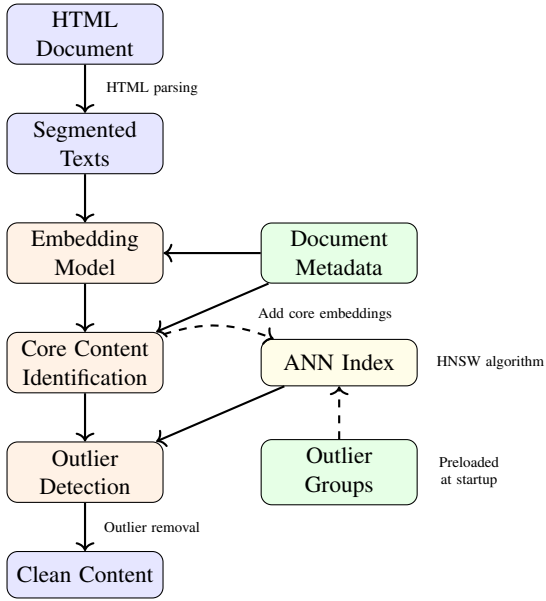

Figure \ref{fig:teaser} illustrates these four steps, showing how segments are embedded, core content is identified, outliers are detected and removed. Figure \ref{fig:architecture} provides an overview of the system architecture, highlighting the key components and data flow.

\subsection{Implementation Optimizations}
For processing millions of documents daily in production, computational efficiency is critical. We optimized SORE through several techniques:

\paragraph{Approximate Nearest Neighbor Search.} 
Computing cosine distances at scale between large numbers of high-dimensional vectors can be computationally expensive. We leveraged Voyager\footnote{\url{https://github.com/spotify/voyager}}, an approximate nearest neighbor (ANN) implementation that uses HNSW (Hierarchical Navigable Small World) under the hood. This provides significant efficiency gains with high accuracy.

\paragraph{Precomputed Indices.} 
During initialization, we create an ANN index and add the outlier group embeddings to it, generating a byte dump of this index. For each document to be cleaned, we load this precomputed index, add the newly computed core content and metadata embeddings, and query for nearest neighbors. This approach avoids rebuilding the entire index for each document.

\paragraph{Optimized Distance Calculations.} 
Since modern embedding models typically produce normalized vectors, we use inner product distance (1 - dot product) rather than full cosine distance computation, reducing computational overhead.

\paragraph{Batched Processing.} 
Embedding computation is performed in batches to maximize throughput when processing multiple documents, optimizing API usage and reducing per-document latency.

In our production Java implementation, the cleanup of each HTML file takes an average of 200 milliseconds, with the external API call for embedding computation accounting for most of the duration (over 100 ms). This performance enables SORE to process millions of documents daily within reasonable time and cost constraints.

\subsection{Key Design Decisions}

\subsubsection{Balancing Efficiency and Semantic Understanding}
SORE addresses three key challenges for industrial deployment: (1) \textbf{Cost efficiency}: LLM inference costs approximately 25× more than our embedding-based approach, saving hundreds of thousands of dollars monthly at scale; (2) \textbf{Latency}: SORE processes documents in 200ms compared to LLMs' 2500ms, meeting strict production constraints; and (3) \textbf{Determinism}: Unlike LLMs that may produce inconsistent results, SORE provides transparent, deterministic explanations for content removal decisions.

\subsubsection{Core Content Identification Strategy}
We chose \textbf{metadata similarity} as our approach for identifying core content, using document metadata as a semantic anchor. This offers several advantages: it typically reflects the document's main theme, is available for most web documents, operates language-agnostically, and establishes a semantic "north star" for identifying relevant content. Empirical testing showed that selecting the top $k\%$ of segments most similar to metadata provides a reliable core content identification mechanism across diverse document types.

\subsubsection{Outlier Group Development}
Our outlier groups were developed through iterative analysis combining data analysis and domain expertise. We implemented \textbf{semantic clustering} to represent outlier groups as clusters in the embedding space, allowing flexible matching of semantically similar content even when exact phrases differ. Each outlier group was tuned through \textbf{precision-recall balancing}, and our production system enables \textbf{continuous refinement} by logging removal decisions for ongoing improvement. The set of outliers used in this study, together with the performance analysis that SORE enables in choosing these keywords, is provided in Appendix \ref{appendix:outliers}.

\section{Experimental Evaluation}

\subsection{Datasets and Evaluation Setup}
We evaluated SORE using two in-house HTML datasets representing real-world content cleaning challenges:

\paragraph{\textsc{SORE-small}} This dataset contains approximately 200 samples with hand-extracted main content from various websites across multiple languages and domains. The manually extracted content serves as a high-quality ground truth for evaluating extraction precision and recall.

\paragraph{\textsc{SORE-large}} This dataset comprises approximately 20,000 samples with automatically extracted ground truth using a combination of ReadabilityJS and n-gram–based content cleanup. It focuses on high precision, removing groups of characters that appear on multiple pages across the web in a multi-million document corpus (e.g., legal disclaimers that appear on every page of a given domain).

For evaluation, we compared SORE against several baseline approaches:

\paragraph{ReadabilityJS} A popular open-source HTML content extractor based on structural heuristics, widely used in industry.

\paragraph{LLM Variants} We tested three LLM-based approaches: (1) LLM (raw HTML) providing the entire HTML content to the LLM for extraction; (2) LLM (raw text) extracting the complete text content from HTML as input; and (3) LLM (tag-depth) a hybrid approach supplying text content along with HTML tag information and tree depth. The relevant LLM prompts and additional discussions can be found in Appendix \ref{appendix:prompts}.

\subsection{Performance Comparison}

\subsubsection{Extraction Quality}
Table \ref{tab:performance} compares SORE against other content extraction methods on \textsc{SORE-small}. SORE achieved a near-LLM level F-score with significantly lower computational requirements.

\begin{table}[!bt]
\centering
\resizebox{\columnwidth}{!}{
\begin{tabular}{lccc}
\toprule
\textbf{Method} & \textbf{F-score} & \textbf{Precision} & \textbf{Recall} \\
\midrule
LLM (tag-depth) & \textbf{0.765} & 0.895 & 0.711 \\
LLM (raw html) & 0.690 & 0.865 & 0.637 \\
LLM (raw text) & 0.583 & 0.795 & 0.520 \\
SORE (cohere, c=0.5, k=10\%) & \textit{0.732} & 0.700 & 0.840 \\
ReadabilityJS & 0.678 & 0.569 & 0.936 \\
\bottomrule
\end{tabular}
}
\caption{Performance comparison on \textsc{SORE-small} across different content extraction methods. SORE achieves near-LLM performance at a fraction of the computational cost. Precision measures the proportion of extracted text that belongs to the ground truth, while recall measures the proportion of ground truth text that was successfully extracted. F-score is the harmonic mean of precision and recall.}
\label{tab:performance}
\vspace{-0.4cm}
\end{table}

The results demonstrate that SORE achieves 96\% of the best LLM approach's F-score (0.732 vs. 0.765) while offering significant advantages in computational efficiency. Notably, SORE outperforms ReadabilityJS by 7.9\% in F-score, with substantially higher precision (0.700 vs. 0.569) while maintaining strong recall.

\subsubsection{Embedding Model Comparison and Parameter Tuning}

\begin{figure}[!tb]
\centering
\includegraphics[width=0.48\textwidth]{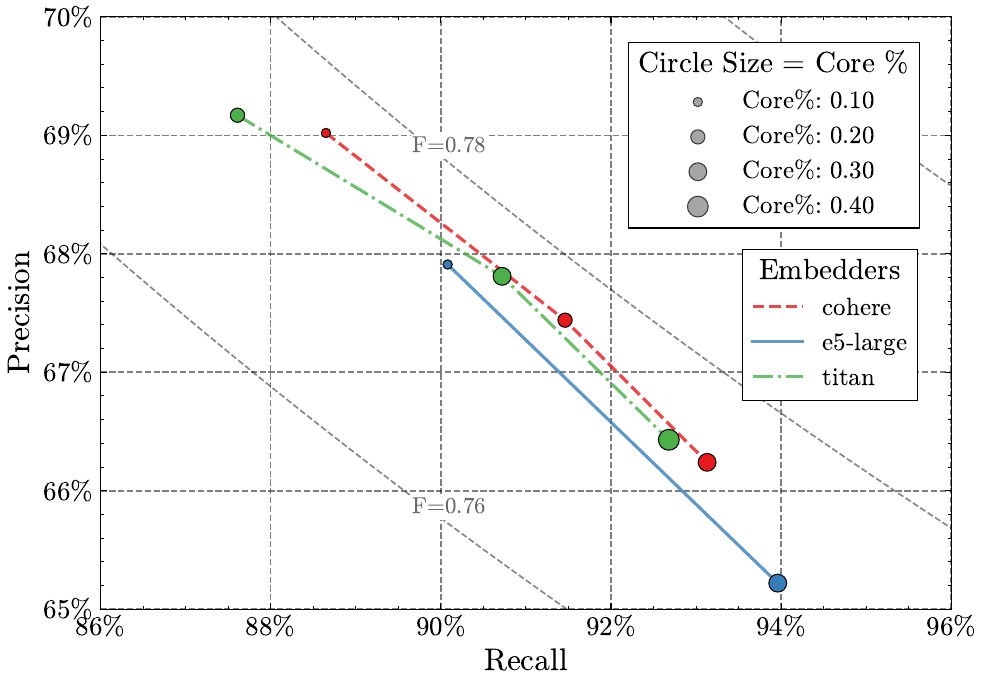}
\caption{Precision-recall trade-offs for different embedding models and SORE parameter settings on \textsc{SORE-large}. AWS Titan (1024d) with core=20\% and cutoff=0.8 provides the best balance of precision, recall, and cost. Each point on the curves represents different parameter configurations.}
\label{fig:precision_recall}
\vspace{-0.4cm}
\end{figure}

Figure \ref{fig:precision_recall} shows the precision-recall trade-offs for various embedding models and parameter configurations on \textsc{SORE-large}. Each point represents a different combination of core percentage ($k$) and embedder type, with the best distance cutoff ($d$) parameters per model family. We compare two commercial solutions, Cohere and AWS Titan, with the open source multilingual embedding model e5-large \cite{wang2024multilingual}.

For this dataset, ReadabilityJS scores 0.596 precision and 0.988 recall, while LLM (tag-depth) achieves 0.885 precision and 0.718 recall (both outside the graph). AWS Titan emerged as the most cost-effective choice ($\sim200$ CHF/month), with comparable performance to more expensive solutions ($\sim1200$ CHF/month for Cohere). The optimal parameters for the AWS Titan-based SORE were found to be 1024-dimensional embeddings, 0.8 distance cutoff, and 0.2 core percentage.

The parameter tuning experiments revealed that higher values of distance cutoff ($d$) increase precision but reduce recall, lower values of core percentage ($k$) make the system more selective but may miss relevant content, and higher-dimensional embeddings generally perform better. These findings enabled us to select parameters that balanced performance and cost for our production deployment.

\subsection{Multilingual Capability and Case Studies}

\subsubsection{Multilingual Performance}
A key advantage of SORE is its language-agnostic operation. Table \ref{tab:examples} presents examples of text segments removed by SORE across multiple languages, demonstrating the system's multilingual capabilities and semantic understanding.

\begin{table*}[!htb]
\centering
\resizebox{\textwidth}{!}{
\begin{tabular}{p{3cm}p{4cm}p{7cm}p{3cm}}
\toprule
\textbf{URL} & \textbf{Title} & \textbf{Removed Text} & \textbf{Reason} \\
\midrule
huffpost.com/... & 10 Things Guests Notice Most About Your Home & SolStock via Getty Images & Source \\
\midrule
foodsguy.com/... & Coconut Sugar Vs Brown Sugar & *This post may contain affiliate links. Please see my disclosure to learn more. & Affiliate Disclosure \\
\midrule
buzzfeed.com/... & This Black Widow Moment... & 03:27 PM - 29 Apr 2019 & Last updated \\
\midrule
dealmoon.com/... & Dyson V12 Detect Slim \begin{CJK*}{UTF8}{gbsn}激光探测无绳吸尘器 翻新\end{CJK*} \$349.99 & \begin{CJK*}{UTF8}{gbsn}点击购买$>$$>$\end{CJK*} & Buy \\
\midrule
blog-rct.com/... & Melvyn Jaminet fait passer un message... & A lire ci-dessous : & Also read \\
\midrule
lapatilla.com/... & & ¡Únete al club ahora! Suscríbete al boletín más importante de Venezuela & Subscribe for free \\
\midrule
cleanmyspace.com/... & Bathroom Cleaning: 10 Things... & Learn More About The 3 Wave Cleaning System & [too irrelevant] \\
\midrule
jagranjosh.com/... & Only People With 20/20 Vision Can Spot... & Your Way Of Clenching Your Fist Reveals Your Hidden Personality Traits & [too irrelevant] \\
\bottomrule
\end{tabular}
}
\caption{Examples of text removed by SORE. The first three rows show examples of removed text with specific reasons. The next three rows demonstrate the system's multilingual capabilities (Chinese, French, Spanish). The last two rows show text removed because it was semantically too distant from the core content.}
\label{tab:examples}
\vspace{-0.4cm}
\end{table*}

Unlike traditional approaches that rely on language-specific patterns or rules, SORE leverages multilingual embedding models that capture semantic relationships across languages. This enables effective content extraction for documents in Chinese, French, Spanish, and other languages without requiring separate models or rule sets.

\section{Industrial Impact and Cost Analysis}

\subsection{Production Deployment}
SORE is currently deployed in a production environment, processing millions of documents daily across multiple languages. The system is implemented as a scalable service that integrates with existing data processing pipelines, providing cleaned content for downstream tasks such as classification and information retrieval.

Our production deployment focuses on four key aspects: (1) \textbf{Horizontal scaling} with multiple instances processing documents in parallel; (2) \textbf{Comprehensive monitoring} capturing performance metrics and removal decisions for continuous improvement; (3) \textbf{Fallback mechanisms} that revert to more conservative extraction when SORE removes unexpectedly large portions of a document; and (4) \textbf{Configurable parameters} that can be adjusted based on specific use cases and language requirements. To promote reproducibility and further research, we will make our implementation and evaluation datasets publicly available.

\subsection{Cost and Efficiency Comparison}
A key advantage of SORE over LLM-based approaches is its significantly lower computational cost. Table \ref{tab:cost} compares the cost and performance characteristics of different approaches.

\begin{table}[!bt]
\centering
\resizebox{\columnwidth}{!}{
\begin{tabular}{lccc}
\toprule
\textbf{Method} & \textbf{F-score} & \textbf{Avg. Latency} & \textbf{Cost per 1M docs} \\
\midrule
LLM (tag-depth) & 0.793 & 2500 ms & \$15,000 \\
ReadabilityJS & 0.743 & 50 ms & \$7 \\
SORE (AWS Titan) & 0.776 & 200 ms & \$600 \\
SORE (Cohere) & 0.777 & 250 ms & \$3,600 \\
\bottomrule
\end{tabular}
}
\caption{Cost and performance comparison using \textsc{SORE-large}. SORE with AWS Titan provides the best balance of performance and cost, with a latency 12.5× lower than LLMs and cost 25× lower per million documents.}
\label{tab:cost}
\vspace{-0.4cm}
\end{table}

SORE achieves near-LLM performance at a fraction of the cost, with 12.5× lower latency (200ms vs. 2500ms) and 25× lower cost (\$600 vs. \$15,000 per million documents) when using AWS Titan embeddings. For our production system processing over 30 million documents monthly, SORE saves approximately \$432,000 annually compared to an LLM-based approach while delivering comparable quality. This substantial cost reduction has made advanced semantic content cleaning viable at scale.

\section{Conclusion}

We introduced SORE (Semantic Outlier Removal), a cost-effective, transparent method for removing unwanted content from web documents while preserving their core message. By leveraging multilingual sentence embeddings and approximate nearest-neighbor search, SORE achieves performance comparable to LLM-based approaches at a fraction of the computational cost.

Our experiments demonstrate that SORE outperforms traditional structure-based methods while maintaining high precision across diverse multilingual scenarios. The system's transparency—providing clear reasons for why specific content is removed—facilitates debugging and continuous improvement.

SORE is currently deployed in production, processing millions of documents daily across multiple languages. Its efficiency and effectiveness make it a practical solution for large-scale content extraction and cleaning in industrial settings. To promote reproducibility and further research in this area, we will make our implementation and evaluation datasets publicly available.

Future work will explore integrating SORE with domain-specific knowledge bases, refining outlier group definitions based on ongoing accuracy analysis, and extending its application to more nuanced tasks such as sentiment-based filtering.

\section*{Ethics Statement}
SORE is designed to extract main content from web pages while respecting copyright and terms of service. The system does not alter the meaning of content but rather removes extraneous elements. We acknowledge the potential risk that in some cases, SORE might remove content that some users consider important. To mitigate this risk, our implementation includes detailed logging of removal reasons and fallback mechanisms when excessive content is removed.

\bibliography{anthology,main}

\clearpage
\appendix
\section{Outlier Groups}
\label{appendix:outliers}

SORE uses a carefully curated set of outlier groups to identify and remove unwanted content. These groups were developed through extensive analysis of web content patterns and iteratively refined based on performance metrics. Each group represents a category of content typically not part of the main article text.

\subsection{Outlier Group Performance}
We analyzed the accuracy of removal for different outlier keywords. Table \ref{tab:keyword_accuracy} shows the least accurate keywords from our analysis.

\begin{table}[!htb]
\centering
\resizebox{\columnwidth}{!}{
\begin{tabular}{lcc}
\toprule
\textbf{Phrase} & \textbf{Occurrence} & \textbf{Accuracy} \\
\midrule
Home & 9777 & 0.510 \\
Frequently asked questions & 822 & 0.540 \\
Similar & 1559 & 0.543 \\
dd/mm/yyyy & 117 & 0.556 \\
Not found & 532 & 0.564 \\
21.02.2023 & 2219 & 0.591 \\
Order & 1996 & 0.599 \\
Error & 2600 & 0.600 \\
URL & 1177 & 0.601 \\
404 & 3391 & 0.602 \\
\bottomrule
\end{tabular}
}
\caption{Removal accuracy for the 10 least accurate outlier keywords. Even the least accurate keywords exhibit accuracy above 0.5, with most outlier groups performing significantly better.}
\label{tab:keyword_accuracy}
\end{table}

The results indicate that some ambiguous terms like "Home" have relatively lower accuracy due to their context-dependent nature—they may appear in both navigation elements and legitimate main content. However, even these challenging outlier groups achieve better than random performance, and the system's overall accuracy benefits from the combination of multiple outlier detection signals.

\subsection{Outlier Group Keywords}

The outlier groups are represented as sets of phrases and patterns that, when embedded, create semantic clusters in the embedding space. The following list shows our production outlier groups organized by category:

\subsubsection{Date-time Related Content}
"Date", "21.02.2023", "21.02.2024", "21.02.2025", "Published at", "Last updated", "Time", "Published", "Updated", "dd/mm/yyyy", "mm/dd/yyyy", "yyyy-mm-dd", "dd.mm.yy"

\subsubsection{Authorship Information}
"Author", "Writer", "Contributor", "Editor", "Posts", "Written by"

\subsubsection{Comment Sections}
"Comment", "Reply", "Feedback", "Discussion", "Leave a comment"

\subsubsection{Source Attribution}
"Source", "Website", "Publisher", "URL", "Link"

\subsubsection{Related Content Links}
"Related", "Read more", "Look:", "Similar", "See also", "Also read", "Read next", "Get more", "Frequently asked questions"

\subsubsection{Calls to Action}
"CTA", "Buy", "Shop", "Order", "Click here", "Check out", "View more", "Visit", "Let me know", "Download", "Subscribe", "Sign up", "Contact us", "Receive notifications"

\subsubsection{Navigation Elements}
"Breadcrumbs", "Home $>$", "Home $>$ About", "Navigation", "Home", "About"

\subsubsection{Contact Information}
"Contact", "Email", "Phone", "Address", "Contact us"

\subsubsection{Social Media Elements}
"Social", "Facebook", "Twitter", "Instagram", "LinkedIn", "TikTok", "Share", "Like", "Follow", "3425 views"

\subsubsection{Legal Content}
"Legal", "Terms", "Privacy", "Policy", "Disclaimer", "Cookie", "Accept", "Policy", "Settings"

\subsubsection{Page Infrastructure}
"Footer", "Copyright", "All rights reserved", "Search", "Find", "Look for", "Explore", "Error", "404", "Not found", "Page not found", "Error", "Try again later"

\subsubsection{Commercial Content}
"Advertisement", "Sponsored", "Promotion", "Sponsor", "Subscription", "Subscribe", "Newsletter", "Membership", "Join", "Affiliate", "Affiliate links", "Disclosure", "Affiliate Disclosure"

\subsubsection{Miscellaneous Boilerplate}
"Refresh this page", "Login required", "License", "Enter your email", "Thank you for reading", "Subscribe for free"

\section{LLM Prompts}
\label{appendix:prompts}

For the LLM baseline comparisons, we systematically developed and tested several prompting strategies. Through empirical evaluation, we found that providing structured context about HTML tags and their depth in the document tree ("tag-depth" approach) yielded the best results, as it strikes a balance between:

\begin{enumerate}
    \item Providing sufficient structural context that pure text approaches lack
    \item Avoiding overwhelming the model with full HTML markup
    \item Creating a constrained output format (line numbers) that prevents hallucination.
\end{enumerate}

The tag-depth approach also significantly outperformed both raw HTML and raw text approaches in our experiments, as shown in Table \ref{tab:performance}. Below are the three prompting strategies we evaluated:

\begin{promptbox}{Raw HTML Prompt}
Analyze the given HTML and extract only the main article/post/discussion content, ensuring that the extracted content meets the criteria for a perfect extraction as defined below.

1. Include All Core Content:
- Extract the complete core content of the main article, which are exclusively:
  - Title
  - Headings and Subheadings
  - All paragraphs that form the continuous, coherent text of the article

2. Exclude All Irrelevant Elements:
- Do not include any peripheral or irrelevant elements such as:
  - Headers, footers, navigation bars, sidebars
  - Comments, author bios, blog names, date stamps, author names, etc.
  - Advertisements (e.g., "Buy now")
  - Breadcrumbs (e.g., "Home $>$ Category $>$ Subcategory")
  - Promotional teasers (e.g., "Sign up for our newsletter")
  - Navigation links (e.g., "Go to the next article")
  - Irrelevant image captions (e.g., "Source: Getty Images")
  - Calls-to-action (e.g., "Join our group")
  - Recommendations for other articles (e.g., "See related article: ...")
  - Contact information (e.g., "Reach us at...")
  - Social media links (e.g., "Connect with @...")
  - Disclaimers or cookie notices

3. Output Format:
- Provide only the main article content without any additional text or commentary.
- Do not include any formatting tags or metadata.

Input HTML:
{text}

Output format:
text
\end{promptbox}

\begin{promptbox}{Raw Text Prompt}
Analyze the given text and extract ONLY the main article content:

1. Identify the core article content, focusing on continuous, coherent text that with a clear title.
2. Ignore all peripheral content: headers, footers, navigation, sidebars, comments, author bios, blog names, date stamps, author names, etc, but do not ignore the content that is included in the main article.
3. Output the main article content.

Input text:
{text}

Output format:
text
\end{promptbox}

\begin{promptbox}{Tag-depth Prompt (Best Performing)}
For the given numbered lines of text from an HTML with their parent tags and the tag depths in the HTML tree, extract the core content (like ReadabilityJS).

1. IDENTIFY CORE CONTENT
- Each page has a main content, which can be an article, blog post, forum thread, etc.
- Extract the main content, which includes the title, headings, paragraphs, and any other relevant text.
- Exclude all peripheral content: headers, footers, navigation, sidebars, comments, author bios, blog names, date stamps, author names, etc.

2. EXCLUDE IF ANY OF THESE ARE TRUE:
- Appears in site navigation sections
- Contains ANY of these patterns:
 * Social media handles or URLs
 * Date stamps or bylines
 * Copyright notices
 * Contact information
 * Newsletter signup text
 * "Related article" references
 * Advertisement markers
 * Image credits or captions
 * Tags or categories
 * Call-to-action phrases
 * Navigation instructions
 * Comment section markers
 * Share button text
 * Footer content
- Some examples are: 'Related: you will not believe what happened next' or 'Sign up to our newsletter' or 'Source: Getty Images' or 'Contact us via Instagram' or 'Date: 2022-01-01'"

3. VALIDATE SELECTION
- Verify selected lines form a coherent narrative
- Check that no essential context is lost
- Confirm removal of ALL peripheral content

Input:
{text}

Output format:
[comma-separated list of line numbers containing ONLY the essential content]

Notes:
- Include ONLY numbers in the output, no explanations
- If a line contains mixed content, exclude it entirely
- When in doubt about a line, exclude it
- Aim for maximum precision over recall

Example output:
1,2,3,5,8,...
\end{promptbox}

\end{document}